\documentclass[10pt,twocolumn,letterpaper]{article}

\usepackage{iccv}
\usepackage{times}
\usepackage{epsfig}
\usepackage{graphicx}
\usepackage{amsmath}
\usepackage{amssymb}
\usepackage{amssymb}
\usepackage{booktabs}
\usepackage{tablefootnote}
\usepackage{dsfont}
\usepackage{algcompatible}
\usepackage{gensymb}
\usepackage{tabularx}
\usepackage{subfigure}
\usepackage[table]{xcolor}  
\usepackage{multicol}
\usepackage{multirow}
\usepackage{makecell}
\usepackage{float}
\usepackage{array}
\usepackage{comment}
\usepackage{caption}
\usepackage{enumitem}

\newcommand{\PreserveBackslash}[1]{\let\temp=\\#1\let\\=\temp}
\newcolumntype{C}[1]{>{\PreserveBackslash\centering}p{#1}}
\newcolumntype{R}[1]{>{\PreserveBackslash\raggedleft}p{#1}}
\newcolumntype{L}[1]{>{\PreserveBackslash\raggedright}p{#1}}

\usepackage[pagebackref=true,breaklinks=true,letterpaper=true,colorlinks,bookmarks=false]{hyperref}
\usepackage[capitalise]{cleveref}
\iccvfinalcopy 

\begin{document}

\title{PointOdyssey: A Large-Scale Synthetic Dataset for Long-Term Point Tracking}

\author{Yang Zheng \quad  
Adam W. Harley \quad
Bokui Shen \quad
Gordon Wetzstein \quad Leonidas J. Guibas \\
Stanford University\\
{\tt\small \{yzheng18,harleya,willshen,gordonwz,guibas\}@stanford.edu}
}

\ificcvfinal\thispagestyle{empty}\fi

\twocolumn[{
\vspace{-10pt}
\maketitle
\begin{center}
    \centering
    \vspace{-2em}
\includegraphics[width=\linewidth]{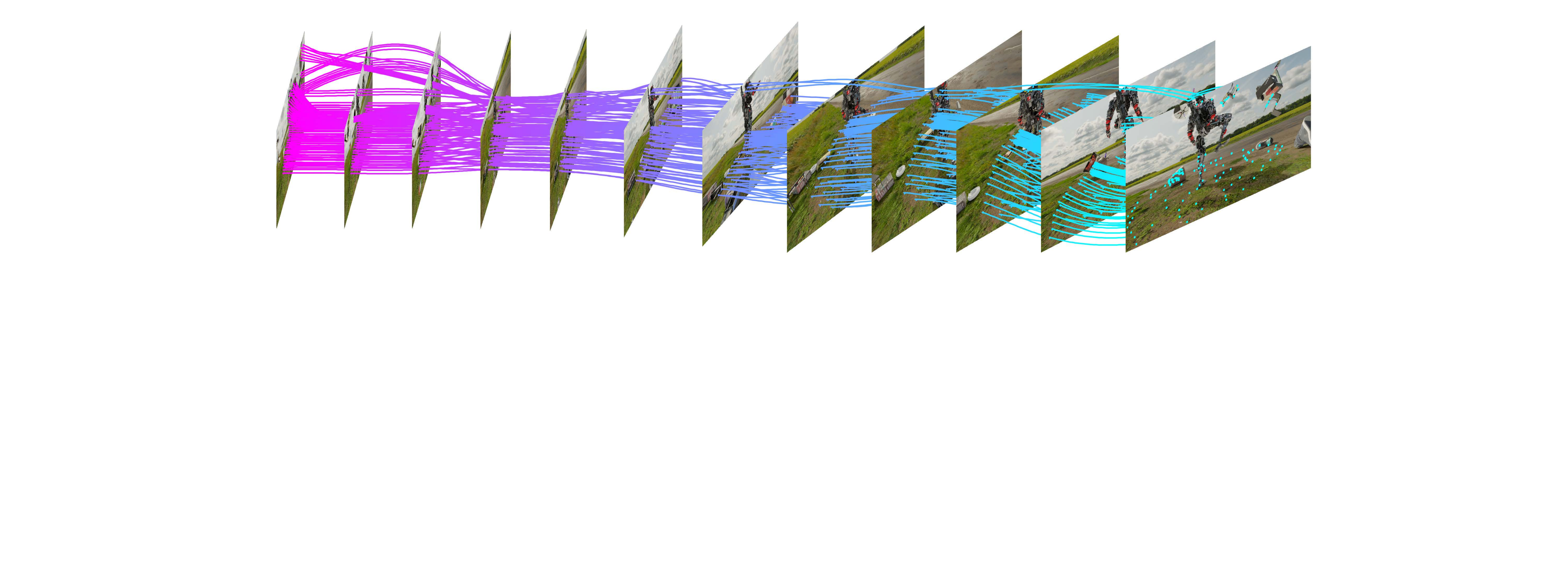}
\captionof{figure}{\textbf{PointOdyssey dataset.} We provide point correspondence annotations across long continuous videos. 
Here we visualize pixel-coordinate trajectories from frame 0 to frame 1600 in a sample video from our dataset.
}
\label{fig:teaser}

\end{center}

}]

\begin{abstract}
\vspace{-0.5em}

We introduce PointOdyssey, a large-scale synthetic dataset, and data generation framework, for the training and evaluation of long-term fine-grained tracking algorithms. Our goal is to advance the state-of-the-art by placing emphasis on long videos with naturalistic motion. Toward the goal of naturalism, we animate deformable characters using real-world motion capture data, we build 3D scenes to match the motion capture environments, and we render camera viewpoints using trajectories mined via structure-from-motion on real videos. We create combinatorial diversity by randomizing character appearance, motion profiles, materials, lighting, 3D assets, and atmospheric effects. Our dataset currently includes 104 videos, averaging 2,000 frames long, with orders of magnitude more correspondence annotations than prior work. We show that existing methods can be trained from scratch in our dataset and outperform the published variants. Finally, we introduce modifications to the PIPs point tracking method, greatly widening its temporal receptive field, which improves its performance on PointOdyssey as well as on two real-world benchmarks. Our data and code are publicly available at: \url{https://pointodyssey.com}

\end{abstract}
\vspace{-0.5em}

\vspace{-0.5em}
\section{Introduction}
\vspace{-0.25em}
In a variety of computer vision tasks, large-scale annotated datasets have provided a highway for the development of accurate models. In this paper, we aim to provide such a highway for the task of \textit{fine-grained long-range tracking}.  
The goal of fine-grained long-range tracking is: given any pixel coordinate in any frame of a video, track the corresponding world surface point for as long as possible.

While there exist multiple generations of datasets targeting fine-grained short-range tracking (\ie, optical flow) \cite{middlebury,sintel,mayer2016large}, and annually updated datasets targeting several forms of coarse-grained long-range tracking (\ie, single-object tracking~\cite{fan2019lasot}, multi-object tracking~\cite{MOTChallenge2015}, video object segmentation~\cite{pont20172017}), there are only a handful of works at the intersection of fine-grained \textit{and} long-range tracking. 

\begin{table*}[t!]
\vspace{-5pt}
\centering
\resizebox{\textwidth}{!}{

\rowcolors{2}{gray!25}{white}

\begin{tabular}{r|C{2.5cm}C{3.5cm}C{2.5cm}C{3.3cm}C{3.1cm}C{2.5cm}}
  \toprule
\multicolumn{1}{l}{} & MPI Sintel~\cite{sintel} & Flyingthings++~\cite{mayer2016large, harley2022particle} & Kubric~\cite{greff2022kubric} & TAP-Vid-Kinetics~\cite{doersch2022tap} & TAP-Vid-DAVIS~\cite{doersch2022tap} & PointOdyssey \\
  \toprule
Resolution & $436 \times 1024$ & $540 \times 960$ & $256\times256$& $\geq 720 \times 1280$ & $1080 \times 1920$ & $540 \times 960$\\
Frame rate & 24 & 8 & 8 & 25 & 25 & 30  \\
Avg. trajectory count & $436\times1024$ & 1,024 & Flexible & 26.3 & 21.7  & 18,700 \\
Avg. span of trajectories & 4\% & 100\% & 100\% & 30\% & 30\% & 100\%  \\
Avg. frames per video & 50 & 8 & 24 & 250 & 67 &  2,035  \\
\midrule 
Training frames & 1064 & 21818 & Flexible & - & - & 166K\\ 
Validation frames & - & 4248 & - & - & -& 24K \\
Test frames & 564 & 2247 & -& 297K & 1999& 26K \\
Total point annotations & 7$\times10^8$ & 3$\times10^{8}$ & -& 8$\times10^7$ & 4$\times10^5$& 4.9$\times10^{10}$ \\
\midrule
Depth \& normals & \checkmark & \checkmark & \checkmark & $\times$ & $\times$ & \checkmark \\
Segmentation masks & \checkmark & \checkmark & \checkmark & $\times$ & $\times$ & \checkmark \\
Retargeted motion & $\times$ & $\times$ & $\times$ & $\times$ &$\times$ & \checkmark \\
Scene randomization &$\times$ & $\times$ & \checkmark & $\times$ & $\times$ & \checkmark \\
Multiple views & $\times$ & $\times$ & $\times$ & $\times$ & $\times$ & \checkmark \\
Continuous & \checkmark & \checkmark & \checkmark & $\times$ & \checkmark & \checkmark \\
Object-object interaction & \checkmark & $\times$ & \checkmark & $\times$ & $\times$ & \checkmark \\
Human-object interaction & \checkmark & $\times$ & $\times$ & \checkmark & \checkmark& \checkmark \\
Human-human interaction & \checkmark & $\times$ & $\times$ & \checkmark & \checkmark& \checkmark \\
\bottomrule
\end{tabular}
}
\caption{
\textbf{Comparison of point tracking datasets.}
PointOdyssey is larger, has longer videos, and includes trajectories which reflect interactions between the objects and the scene. Note that the TAP-Vid datasets are real-world, with sparse human annotations, and are typically reserved for testing~\cite{doersch2022tap}, whereas most synthetic datasets provide train/test splits. 
}
\label{tab:stat}
\vspace{-1em}
\end{table*}

Harley~\etal~\cite{harley2022particle} and Doersch~\etal~\cite{doersch2022tap}, train fine-grained trackers on unrealistic synthetic data (FlyingThings++~\cite{mayer2016large,harley2022particle} and Kubric-MOVi-E~\cite{greff2022kubric}), consisting of random objects moving in random directions on random backgrounds, and test on real-world videos with sparse human-provided annotations (BADJA~\cite{badja} and TAP-Vid~\cite{doersch2022tap}). 
While it is interesting that generalization to real video emerges from these models, the use of such simplistic training data precludes the learning of long-range temporal context, and scene-level semantic awareness. We argue that long-range point tracking should not be treated as an extension of optical flow, where naturalism might indeed be discarded without ill effect~\cite{sun2021autoflow}. Pixels in real video may move somewhat unpredictably, but they take a journey which reflects a variety of modellable factors, including camera shake, object-level motions and deformations, and multi-object relationships such as physical and social interactions. Realizing the grand scope of this problem, both in our data and in our methods, is critical for progress. 

We propose \textit{PointOdyssey}, a large-scale synthetic dataset for the training and evaluation of long-term fine-grained tracking. Our dataset aims to provide the complexity, diversity, and naturalism of real-world video, with pixel-perfect annotation only possible in simulation. 
Besides the length of our videos, the key aspects differentiating our work from prior synthetic datasets are (1) we use motions, scene layouts, and camera trajectories mined from real-world videos and motion captures (as opposed to being random or hand-designed), and (2) we use domain randomization on a wider range of scene attributes, including environment maps, lighting, human and animal bodies, camera trajectories, and materials (similar to Shen~\etal~\cite{shen2021igibson}). Thanks to progress in the availability of high-quality assets and rendering tools, 
we are also able to deliver better photo-realism than possible in years past. 

The motion profiles in our data come from large-scale motion-capture datasets of humans and animals~\cite{mahmood2019amass,li20214dcomplete}. We use these captures to drive humanoids and animals in outdoor scenes, producing realistic long-range trajectories. In outdoor scenes, we pair these actors with 3D assets randomly scattered on the ground plane, which react to the actors according to physics (\eg, being kicked away as the feet collide with the objects). To produce realistic indoor scenes, we use motion captures of indoor scenes~\cite{zheng2022gimo, zhang2022egobody}, and \textit{manually replicate} the capture environments in our simulator, allowing us to re-render the exact motions and interactions, and preserve their scene-aware nature. 
Finally, we import camera trajectories computed from real video~\cite{li2019learning}, and attach additional cameras to the synthetic humans' heads, giving challenging multi-view data of the scenes. 
Our capture-driven approach is in contrast to the mostly random motion patterns used in Kubric~\cite{greff2022kubric} and FlyingThings~\cite{mayer2016large}.
We hope that our data will encourage the development of tracking methods which use scene-level cues to provide strong priors on tracking, pushing past the tradition of relying entirely on bottom-up cues such as feature-matching. 

Our data's visual diversity stems from a large set of simulated assets: $42$ humanoid shapes with artist-made textures, $7$ animals, $1$K+ object/background textures, $1$K+ objects, $20$ unique 3D scenes, and $50$ environment maps. 
We randomize the scene lighting to achieve a wide range of dark and bright scenes. We also render dynamic fog and smoke effects into our scenes, introducing a form of partial occlusion entirely missing from FlyingThings and Kubric. 

PointOdyssey unlocks a variety of new challenges, one of them being: how to use long-range temporal context. Since prior datasets have only included short videos for training ($<30$ frames, see Table~\ref{tab:stat}), existing models only exploit similarly short temporal context. For example, the current state-of-the-art method Persistent Independent Particles (PIPs)~\cite{harley2022particle}, uses an 8-frame temporal window when tracking. As a step toward leveraging \textit{arbitrarily long} temporal context, we propose some modifications to PIPs~\cite{harley2022particle}, greatly widening its 8-frame temporal window, and incorporating a template-update mechanism. Experimental results show that our method achieves higher tracking accuracy than all existing methods, both on the PointOdyssey test set and on real-world benchmarks.

In summary, the main contribution of this paper is \textit{PointOdyssey}, a large-scale synthetic dataset for long-term point tracking, which aims to reflect the challenges---and opportunities---of real-world fine-grained tracking. The dataset, and the code for the simulation engine, are available at: \url{https://pointodyssey.com}
\section{Related  Work}\label{sec:related}
\vspace{-0.25em}

\noindent\textbf{Motion Datasets.}
For many years, the Middlebury dataset~\cite{middlebury} was the primary benchmark for stereo and motion estimation methods. This dataset contains a mix of synthetic and real data, with high-quality annotations, but is a very small dataset by today's standards ($<100$ frames). The MPI Sintel dataset~\cite{sintel} provided a large step forward in visual and motion diversity, by extracting $1064$ frames from a movie animated in Blender~\cite{blender}, including lighting variation, shadows, specular reflections, complex materials, and atmospheric effects. Our dataset is similar to Sintel, but is orders of magnitude larger, both in overall frame count and in the length of the video clips, and is also far more realistic, making use of rendering advancements in Blender. 

The KITTI dataset~\cite{kitti} provides stereo and flow annotations for real-world driving scenes. Real-world annotation is difficult, and therefore approximated: the authors use LiDAR combined with egomotion information to estimate motion in the static parts of the scene, and then fit 3D models to the cars to estimate the motion of car pixels. We opt for synthetic data generation to avoid these approximations and to ensure perfect fine-grained ground truth. 

A series of synthetic datasets have been introduced specifically for training neural nets for motion estimation: FlyingChairs \cite{flownet}, FlyingThings3D \cite{mayer2016large}, FlyingThings++ \cite{harley2022particle}, AutoFlow~\cite{sun2021autoflow}, and Kubric~\cite{greff2022kubric}. These datasets consist of random objects moving in random directions on random backgrounds, yielding unrealistic but extremely diverse data. Of these, only FlyingThings++~\cite{harley2022particle} and Kubric-MOVi-E~\cite{greff2022kubric} provide multi-frame trajectories (as opposed to 2-frame motion). Our dataset has similar motivations, in terms of enabling generalization via diversity, but is targeted toward longer-range tracking---across thousands of frames, instead of merely dozens. Our dataset also includes humans, which interact with each other and with the scene, which we hope will give advantage to methods that use high-level contextual cues (such as scene layout), in addition to the low-level motion and appearance signals.

The recently released TAP-Vid benchmark~\cite{doersch2022tap} aligns well with our work: it argues for the importance of fine-grained multi-frame tracking, and suggests a train/test pipeline where training happens in synthetic data (Kubric-MOVi-E~\cite{greff2022kubric} and RGB-stacking~\cite{lee2021beyond}), and testing happens in real data, which consists of manually annotated point tracks for videos in Kinetics~\cite{kay2017kinetics} and DAVIS~\cite{davis2017}. We show that by training in our new richer synthetic data, we improve performance on the TAP-Vid test set. The ``test'' split of our dataset also covers a gap in TAP-Vid by providing accurate annotations \textit{during occlusions}, while TAP-Vid only provides annotations during visibility. 

There is also a long line of work which trains directly on unlabelled data, using a variety of auxiliary objectives to encourage tracking to emerge~\cite{yu2016back,vondrick2018tracking,wang2019learning,jabri2020walk,bian2022learning}.
An advantage of these works is that they need not worry about a sim-to-real gap, because they train directly on real video. On the other hand, current rendering tools deliver such high photo-realism that the risk of a sim-to-real gap may be much smaller than seen in years past, making synthetic supervision increasingly viable~\cite{Dosovitskiy17,wang2020tartanair}. 

\vspace{1mm}\noindent\textbf{Motion Understanding.}
Early motion estimation methods cast point tracking as an optimization problem defined on handcrafted features \cite{Horn:1981,lucas1981iterative,tomasi1991detection,brox_densepoint,Brox2011LargeDO}, and these techniques continue to drive structure-from-motion \cite{bregler_recover,kong_deepnrsfm,novotny2019c3dpo} and simultaneous localization and mapping systems \cite{taketomi2017visual}. Given the success of neural networks in other computer vision tasks, researchers now typically train deep neural nets to solve the task in a feedforward manner~\cite{flownet,flownet2}, or mix feedforward and iterative inference~\cite{sun2018pwc,teed2020raft}. 

While most early work focuses on estimating optical flow (the motion field that links two consecutive frames), there has recently been a push to estimate fine-grained correspondences across multiple frames. PIPs~\cite{harley2022particle} estimates 8-frame trajectories for pixels, using a learned iterative inference procedure that considers match costs and an implicit temporal prior, considering all 8 timesteps jointly with a powerful MLP-Mixer~\cite{tolstikhin2021mlp}. These 8-frame trajectories can be chained across time to produce longer-range tracks, but these longer tracks are more susceptible to drift, and slow to compute. TAP-Net~\cite{doersch2022tap} estimates correspondences for pixels by taking the argmax of frame-by-frame cost maps, which are computed efficiently using time-shifted convolutions~\cite{lin2019tsm}. 
Empirically, TAP-Net outperforms PIPs when there are long occlusions or hard cuts in the video, likely because the 8-frame temporal window in PIPs is incapable of resolving occlusions that exceed this window, and because hard cuts are inconsistent with the learned prior~\cite{doersch2022tap}. 

\begin{figure*}[ht!] 
    \vspace{-0.5em}
    \centering
    \includegraphics[width=\linewidth]{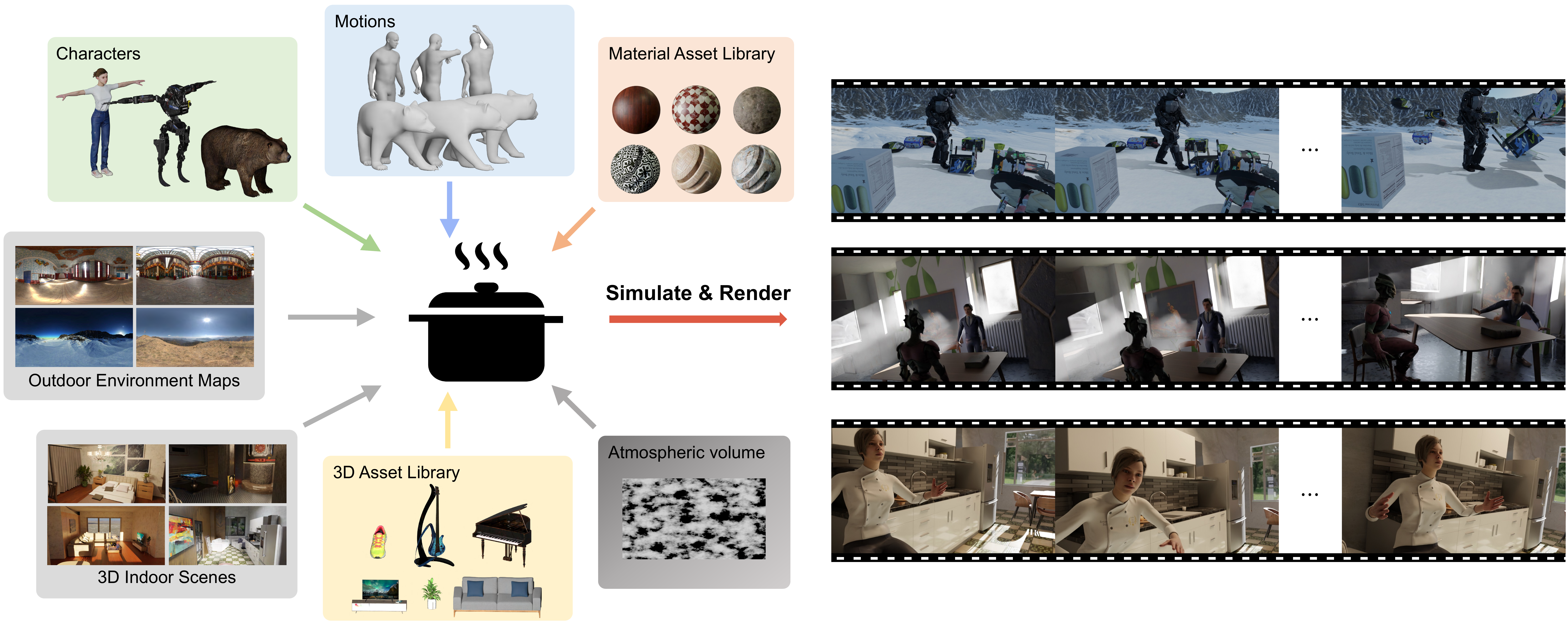}
    \vspace{-17pt}
    \caption{\textbf{Overview of our data generation pipeline.} We randomly generate physically realistic and semantically plausible scenes, by sampling human and animal subjects, motion trajectories for the subjects and the camera, 3D physical assets, materials, environment maps for outdoor scenes, manually created environments for indoor scenes, as well as lighting and atmospheric effects.  
    From these scenes we render videos, paired with various ground truth.
    }
    \vspace{-1em}
    \label{fig:dataset}
\end{figure*}

In this work, we extend PIPs by eliminating its hard 8-frame constraint, allowing it to take much wider temporal context into account. We achieve this by replacing the MLP-Mixer component (in which some parameters were tied to the size of the temporal window), with a deep 1D convolutional network (in which fixed-length kernels are applied to arbitrary temporal spans). We show that our model, trained from scratch in PointOdyssey, 
outperforms both PIPs and TAP-Net. Additionally, we retain a key advantage of PIPs over TAP-Net, which is the ability to produce reasonable estimates \textit{during occlusions}, by tracking multiple timesteps jointly instead of frame-by-frame.

\section{PointOdyssey Dataset}
\vspace{-0.25em}
A sample from our dataset is shown in~\cref{fig:teaser}, and an overview of our data generation pipeline is shown in \cref{fig:dataset}. To generate complex but realistic long-range motion, we use humanoids, robots, and animals, driven by motion capture data~\cite{mahmood2019amass, li20214dcomplete, zheng2022gimo, zhang2022egobody}. This allows us to render long-term dynamic sequences that incorporate long-range interactions between the deformable characters and the 3D environments. We maximize the diversity of the dataset by randomizing the scenes with various materials, textures, and lighting. To add further visual complexity, we introduce random noise to the scene volume density to create changing fog and smoke, which act as a natural occluder and have a significant impact on the appearance and visibility of the scene. This section summarizes our data collection process.

\subsection{Long-Term Motion Data}
\vspace{-0.25em}

\noindent\textbf{Deformable Characters.} We collect 42 open-sourced artist-designed humans and robots from BlenderKit~\cite{blenderkit}, Mixamo~\cite{Mixamo}, and TurboSquid~\cite{turbosquid}, along with 7 animals from DeformingThings4D~\cite{li20214dcomplete}. These assets provide high-poly meshes, along with photorealistic materials and textures, and are rigged to enable animation. 

\vspace{0.5em}\noindent\textbf{Motion Retargeting.} To animate the humanoid characters, we use real-world human motion data~\cite{mahmood2019amass, zheng2022gimo, zhang2022egobody}. We retarget the source motions represented as SMPL-X~\cite{SMPL-X:2019} sequences to target characters, using the motion retargeting algorithm from the Rokoko Toolkit~\cite{Rokoko}. Defining $S_{Rig}$ as the rig of the SMPL-X human model and $T_{Rig}$ as the rig of the target character with $z$-axis up in the resting body pose, we equalize the scale between two rigs as:
\begin{equation}
s=\frac{Z_{max}(T_{Rig})-Z_{min}(T_{Rig})}{Z_{max}(S_{Rig})-Z_{min}(S_{Rig})}\,,
\end{equation}
where $s$ is then applied to the source rig as $S'_{Rig}=S_{Rig}/s$.
Defining $B_i(p_i)$ as a bone in a rig parameterized by $p_i$, (\eg, head and tail location and rotation) we align the source rig with the target rig, setting $p_i^{S^*}=p_j^{T}$, where $p_i^{S^*}$ is the parameter of the bone $B_i$ in the source rig, and $p_j^{T}$ denotes the parameter of the corresponding bone in the target rig, using the bone mapping between the two rigs. Using the aligned source rig $S^*_{Rig}$, we copy the animation to the target rig. 

\begin{figure}[t!] 
    \centering
    \includegraphics[width=\linewidth]{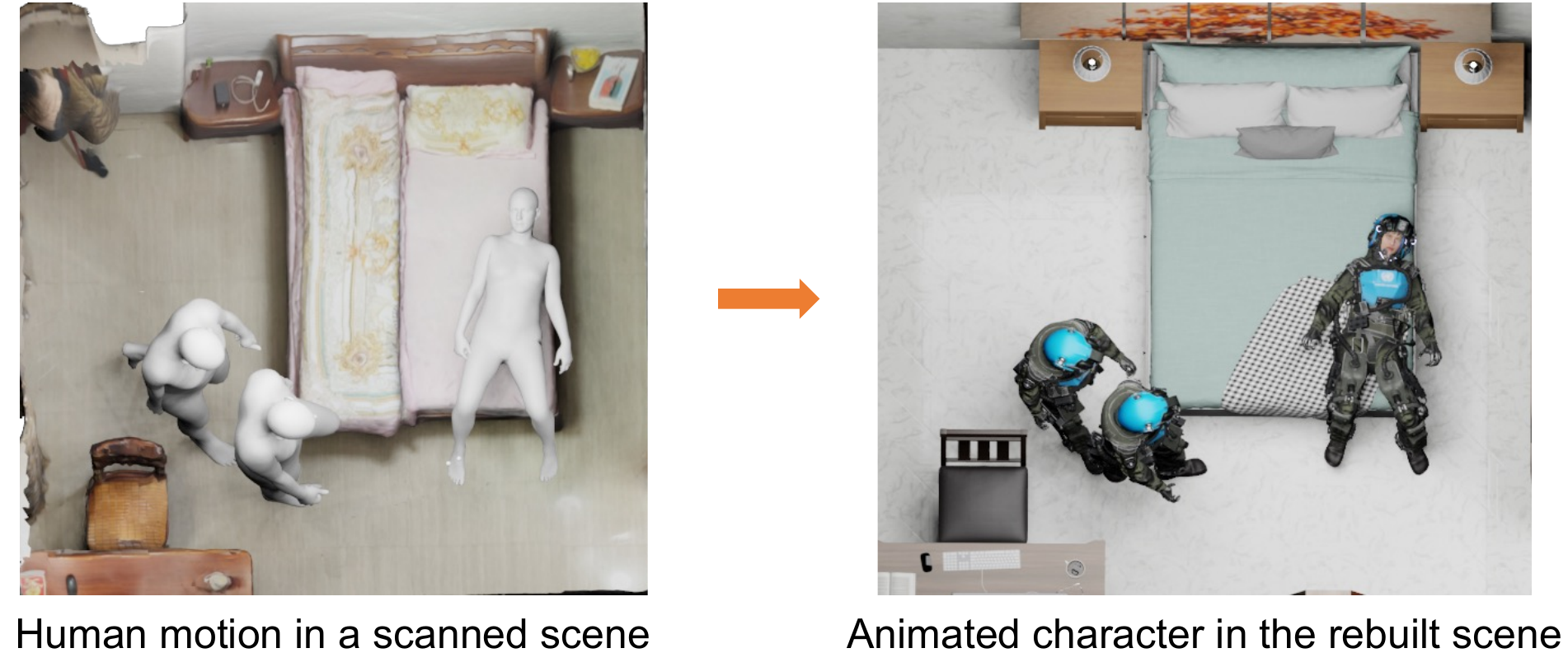}
    \vspace{-1.5em}
    \caption{We use real-world motion capture data~\cite{zheng2022gimo, zhang2022egobody} within 3D scenes manually re-built to match the motion capture environments.
    }
    \label{fig:rebuilt}
    \vspace{-1.5em}
\end{figure}
For animals, since the motions in DeformingThings4D~\cite{li20214dcomplete} are already bound to the meshes, we do not need a retargeting process.

\subsection{3D Environment Context}
\vspace{-0.25em}
Our dataset contains outdoor scenes, which involve randomized but physically coherent agent-object and object-object interactions, and indoor scenes, which involve realistic agent-scene and agent-agent interactions.

\vspace{0.5em}\noindent\textbf{Outdoor Scenes.} Similar to Kubric~\cite{greff2022kubric}, we populate outdoor scenes with random rigid objects from GSO~\cite{downs2022google} and PartNet~\cite{mo2019partnet}. We animate our deformable characters to move around in these scenes, and treat these characters as passive objects with infinite mass, causing the scattered rigid objects to react as though being kicked. We also apply random forces to the rigid objects at random timesteps, to create difficult near-random motion trajectories, with realistic physical collisions. We use HDR environment textures collected from PolyHaven\cite{Polyhaven} mapped into a dome-like region~\cite{LilySurfaceScraper} to simulate natural backgrounds.

\vspace{0.5em}\noindent\textbf{Indoor Scenes.}
We manually build twenty 3D indoor scenes to replicate specific 3D environments from our motion capture datasets~\cite{zheng2022gimo, zhang2022egobody}, matching the scene layouts and furniture as closely as possible, sourcing furniture assets from Blenderkit~\cite{blenderkit} and 3D-FRONT~\cite{fu20213d, fu20213dfuture}. 
We then use motion capture data from these same scenes to animate our characters in the environments, yielding collision-free and naturalistic motion, as shown in~\cref{fig:rebuilt}. We note that unlike the outdoor scenes and unlike prior work, these motions reflect true affordances of the 3D environments. 

\begin{figure}[t!] 
    \centering
    \vspace{-7pt}
    \includegraphics[width=\linewidth]{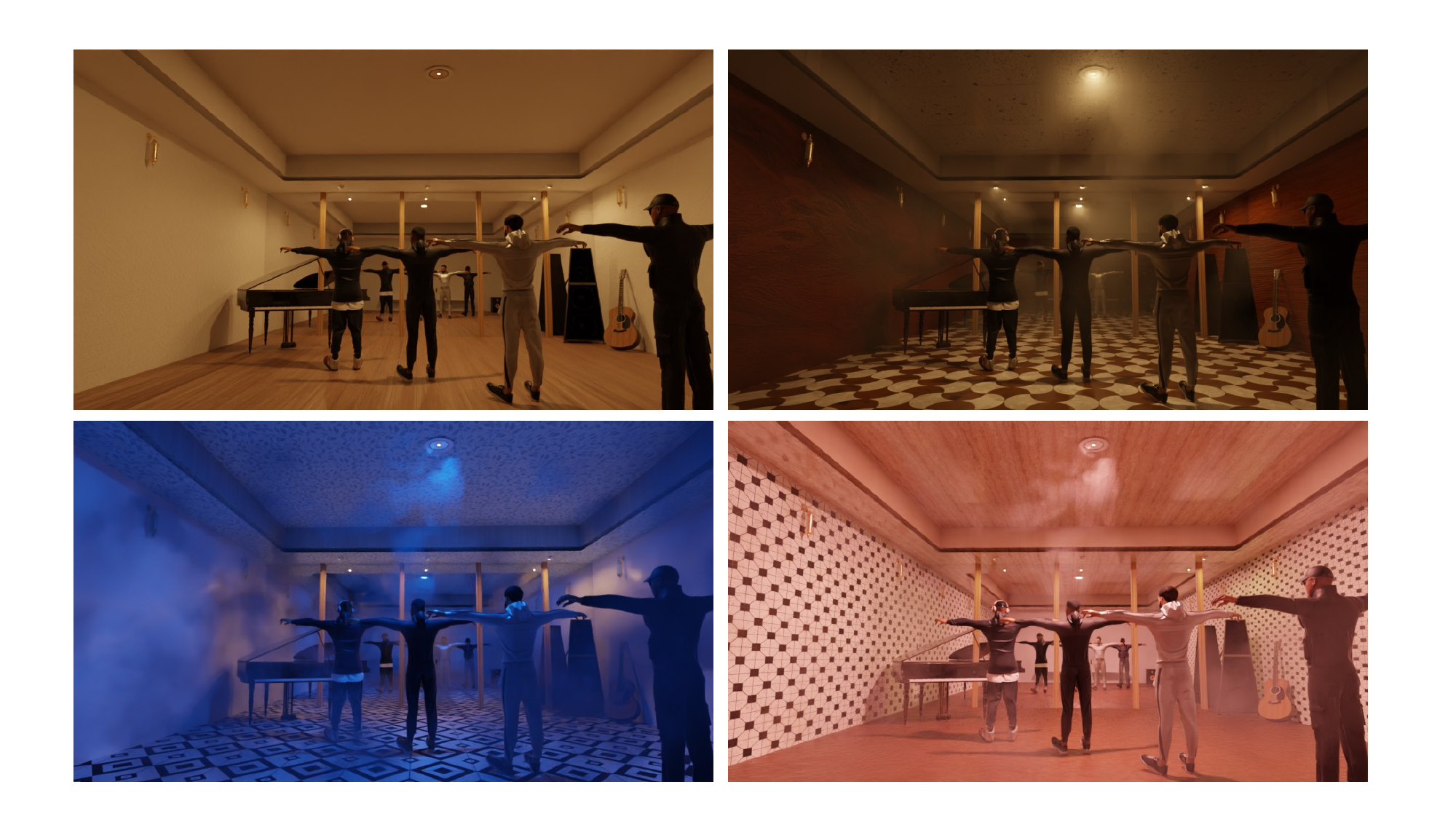}
    \vspace{-15pt}
    \caption{We randomize the textures, materials, atmospheric volume, and lighting, to maximize data diversity.} 
    \vspace{-15pt}
    \label{fig:random}
\end{figure}

\subsection{Camera Motion}
\vspace{-0.25em}
For outdoor scenes, we drive the camera using trajectories extracted from YouTube videos via structure-from-motion~\cite{li2019learning}. 
For indoor scenes, we manually create cinematic camera trajectories consisting of orbits, swoops, and zooms, as well as render ego-centric videos by attaching cameras to the heads of the virtual subjects. Similar to real-world egocentric video~\cite{damen2020epic}, our synthetic ego-centric views yield particularly challenging motion trajectories. 

\subsection{Scene Randomization}
\vspace{-0.25em}
We add diveristy by randomizing our synthetic scenes, in steps similar to iGibson~\cite{shen2021igibson}. For indoor scenes, we randomize the texture of floors, walls, and ceilings, by sampling from $80$ high-quality materials from BlenderKit~\cite{blenderkit}, and randomize the lighting. For outdoor scenes, we randomize the textures of objects by sampling from $1000$ texture maps from GSO~\cite{downs2022google}; we randomize the appearance of the animals by sampling from $24$ high-fidelity fur materials from Blenderkit; we randomize the background by sampling from $50$ $4$K-resolution HDR images from PolyHaven~\cite{Polyhaven}. We additionally generate fog and smoke by adding procedural atmospheric effects to the scene volume. As shown in~\cref{fig:random}, these scene randomization steps add diversity and difficulty to the data.

\subsection{Annotation Generation}
\begin{figure}[t!] 
    \centering
    \vspace{-7pt}
    \includegraphics[width=\linewidth]{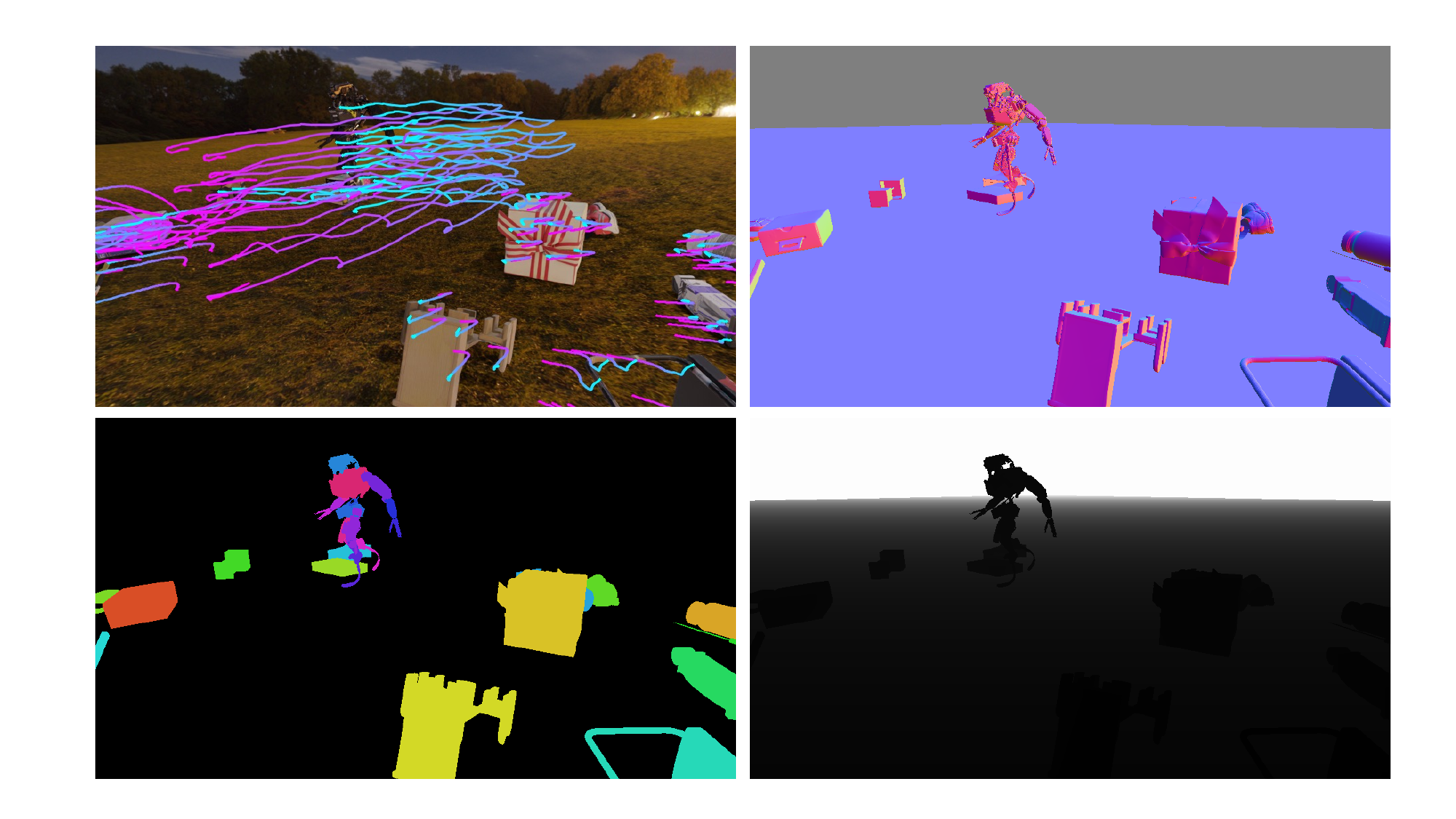}
    \vspace{-15pt}
    \caption{We export 2D and 3D point trajectories, instance masks, depth, normals, and camera calibration data.} 
    \vspace{-15pt}
    \label{fig:anno}
\end{figure}

\vspace{-0.25em}
We generate point trajectories by exporting tracked 2D and 3D coordinates of random foreground and background vertices. We additionally compute visibility annotations, by comparing the depth of the tracked points to the rendered depth values at the projected coordinates. As shown in~\cref{fig:anno}, we also export depth, normals, instance segmentation, camera extrinsics, and camera intrinsics. While our focus is on point tracking, we hope these extra annotations will support a wide set of applications.

\subsection{Statistics}
\vspace{-0.25em}
Our dataset consists of $43$ outdoor scenes and $61$ indoor scenes, totaling $216$K $540\times960$ images at $30$ FPS. The data was rendered in ~$2600$ GPU hours using the Cycles engine in Blender. We divide the dataset into $166$K frames for training, $24$K frames for validation, and $26$K frames for testing. \cref{tab:stat} summarizes key statistics of our dataset compared to related works.

\section{Long-Term Tracking with PIPs++}
\vspace{-0.25em}

In this section, we propose a method that takes advantage of PointOdyssey's realistic long-range motion annotations, both to establish a reasonable benchmark on the dataset's ``test'' split, as well as to improve state-of-the-art on real-world performance.
We base our approach on ``Persistent Independent Particles'' (PIPs)~\cite{harley2022particle}, a state-of-the-art method for fine-grained tracking. 
Its main advantage over prior work is that it inspects $8$ frames at a time, whereas prior work typically used just $2$.
This gives the model some robustness to occlusions, since it can use frames before and after occlusions to estimate the missing parts of the trajectory. We highlight two key limitations, which we aim to address in the following subsections: (1) the temporal field of view is \textit{only} $8$ frames, meaning that the method cannot survive occlusions which are longer than this timespan, and (2) the model relies entirely on the first-frame appearance of the target, making correspondence difficult across appearance changes.
We begin by describing the PIPs architecture in detail, and then describe how we resolve these limitations. 

\subsection{Preliminaries (PIPs)}
\vspace{-0.25em}

PIPs takes an $8$-frame RGB video as input, along with a coordinate $p_1=(x_1, y_1)$ indicating a target to track. It produces a $8 \times 2$ matrix as output, representing the trajectory of the target across the given frames. This process can be repeated across $8$-frame segments, to produce long-range tracks. An arbitrary number of targets can be tracked in parallel, but there is no message-passing between the trajectories (hence persistent \textit{independent} particles). Inference has two main stages: initialization, and an iterative loop. 

\vspace{0.5em}\noindent\textbf{Initialization.} 
Before tracking begins, we compute a feature map $\mathcal{F}_t$ for each frame, with a 2D residual convnet~\cite{he2016deep}. We obtain a vector representing the appearance of the target, by bilinear sampling at the target's position on the first frame's feature map: $f_{p_1}=\texttt{sample}(\mathcal{F}_t, p_1)$. Using this first coordinate and feature vector, we initialize a list of positions and features, $\{(p_t, f_{t})\} = \{(p_1, f_{1})\}$ for all $t\in\{1, 2, \cdots, T\}$.

\begin{figure}[t!] 
    \centering
    \includegraphics[width=\linewidth]{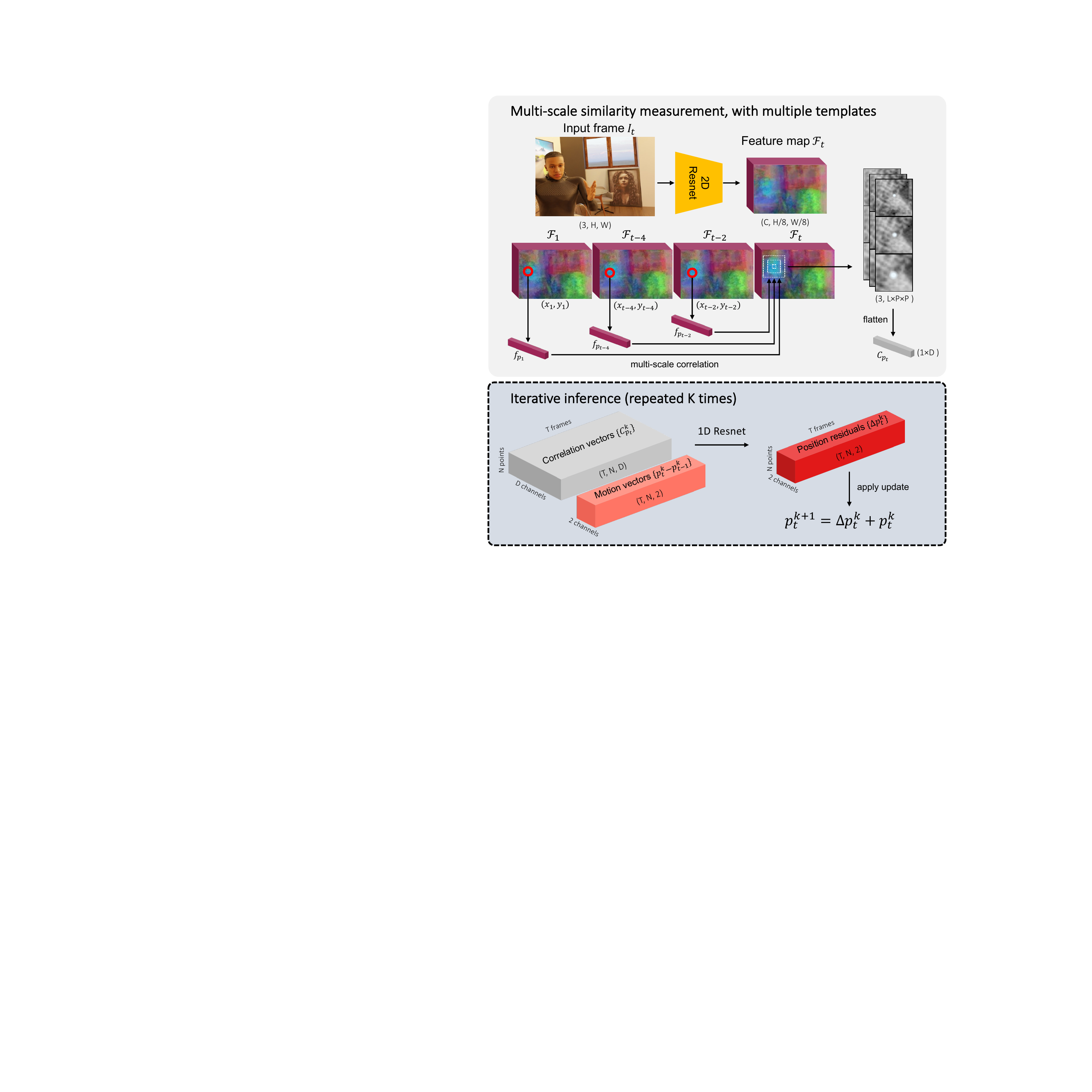}
    \vspace{-15pt}
    \caption{Overview of our method. {Top}: for any query point $p_t$, we first compute the similarity cost volume $C_{p^t}$. We propose to leverage informative features within the local context and incorporate global guidance to maintain consistent and robust tracking. {Bottom}: we iteratively update the trajectory of positions with a 1D Resnet.} 
    \vspace{-15pt}
    \label{fig:network}
\end{figure}

\vspace{0.5em}\noindent\textbf{Iterative Updates.} 
The main inference stage is an iterative update process, which primarily aims to improve the positions $p_t$, so that they track the target more closely. Denoting the current iteration's workspace on iteration $k$ as $\{(p_t^k, f_{t}^k)\}$, we begin an iteration by measuring the similarity between the per-timestep feature vectors and the per-timestep feature maps, within local windows centered at the current estimates: 
\begin{equation}
    C^k_{p_t}=f_{t}^k\otimes \texttt{multicrop}(\mathcal{F}_{t}, p_t^k)/\sigma\,
    \label{eq:corr}
\end{equation}
where $\otimes$ denotes a dot product, $\texttt{multicrop}(\mathcal{F}_{t}, p_t^k)$ produces multi-scale crops from $\mathcal{F}_{t}$ centered at $p_t^k$, and $\sigma$ is a temperature parameter. A 12-block MLP-Mixer~\cite{tolstikhin2021mlp} takes these correlations as input, along with the apparent point motions $p_t^k-p_1^k$, and the features $f_{t}$, and produces updates to the full sequence of positions and features: $\{\Delta p_t^k, \Delta f_t^k\}$. 
These updates are then applied additively, which leads to sampling new local correlations in the next iteration. The feature vectors are eventually fed to a linear layer, which produces per-timestep visibility estimates. 

\vspace{0.5em}\noindent\textbf{Limitations.} PIPs is locked to the temporal field of view that it is trained with, due to the use of the MLP-Mixer in the iterative stage. While the tracker can be chained across time to produce long tracks, these are sensitive to drift, especially when the target becomes occluded beyond the range of the temporal window. We also note that the \textit{visibility-aware} chaining proposed in PIPs cannot be easily parallelized, and so long-range multi-particle tracking is computationally very expensive.  Additionally, we point out that the feature-update operator \textit{cannot} perform a task resembling a template-update, because it does not have access to the input frames. The residual updates to the feature list likely only serve visibility estimation. 

\subsection{Expanding the temporal field of view (PIPs+)}\label{sec:1dres}
\vspace{-0.25em}

Our first proposed modification to PIPs aims to widen its temporal field of view, and enable longer-range tracking. The key component here is the MLP-Mixer, which (by design) has a fixed-width temporal field of view, set to $8$ in PIPs. We propose to replace the MLP-Mixer with an 8-block 1D Resnet~\cite{he2016deep}, doing convolutions across time.\footnote{In the PIPs paper, Harley et al.~\cite{harley2022particle} briefly mention an unsuccessful attempt at using temporal convolutions instead of the MLP-Mixer. It may be that their effort failed due to lack of long-sequence training data.} This means learning kernels that slide across the time axis. Each residual block consists of two convolution layers with kernel size 3, with instance normalization~\cite{ulyanov2016instance} and ReLU~\cite{agarap2018deep}. At the final block, the receptive field is 35 timesteps. Note however that since this module is \textit{iterated} during inference, the effective receptive field is much larger.  

We find that this convolutional variant of PIPs, which we name PIPs+, improves long-range tracking accuracy, and also speeds inference in long videos (from $4$ FPS on average, to $55$ FPS on average, at $720 \times 1080$ on an Nvidia V100 GPU). The convolutional  design enables us to train and test with videos of different lengths, similar to how fully convolutional 2D networks can train and test with different image sizes, but in practice we find it is still important to train and test with roughly similar sequence lengths.

\subsection{Extending to multiple templates (PIPs++)}
\label{sec:multi_app}
\vspace{-0.25em}

Tracked targets are likely to undergo appearance changes across time, and it is important to keep up with these changes. In the original PIPs architecture, the first-frame feature $f_{1}$  (ignoring the negligible feature-update step already discussed) is used for cross-correlation on every frame in the temporal span. This is liable to produce weak matches after appearance changes, and erroneous matches during occlusions. Our second proposed modification to PIPs aims to tackle this ``template update'' problem~\cite{matthews2004template}.

Our main idea is simply to accommodate appearance changes by collecting ``recent appearance'' templates along the estimated trajectory, to complement the ``initial appearance'' template from the first frame. 
Specifically, when computing local correlations for frame $t$, we use the estimated trajectory to extract new features at fixed temporal offsets from this timestep, such as $\{t-2, t-4\}$. This means using $p_{t-k}$ to extract a temporary feature vector $f_{t-k} = \texttt{sample}(\mathcal{F}_{t-k}, p_{t-k})$. We use these features to compute additional correlations in the current frame's feature map $\mathcal{F}_t$, as done in Eq.~\ref{eq:corr}. This process is illustrated in~\cref{fig:network}. The key idea is that if tracking was successful on one of these offset frames, then the extracted feature will reflect the updated appearance of the target, and will yield a more-useful correlation map than the one from $f_1$. These multiple correlation maps are simply concatenated, increasing the channels input to our 1D Resnet. Note that similar to current methods in object tracking~\cite{yan2021learning}, we always retain the initial template $f_1$, to help prevent ``forgetting''.

An alternative strategy here would be to generate templates exclusively from timesteps with high visibility confidence, as commonly done in object tracking~\cite{yan2021learning}. While this is intuitively appealing, we note that our simpler strategy instead allows the model to (temporarily) capture the appearance of an \textit{occluder}, which can sometimes be the appropriate entity to track (\eg, during object self-occlusions). 

Our multi-template strategy, combined with temporally-flexible computation, obviates the residual feature updates, so we simply omit this component. We also omit visibility estimation for simplicity. We name our full model PIPs++.

\section{Experiments}
\vspace{-0.25em}

In this section we explain our experimental setup and results. We recommend watching the supplementary video for better visualization of our dataset and results. 

\subsection{Experimental setup}
\vspace{-0.25em}

\noindent\textbf{Baselines.} We benchmark point trackers, PIPs~\cite{harley2022particle}, TAP-Net~\cite{doersch2022tap}, our proposed PIPs+ and PIPS++, an optical flow method 
RAFT~\cite{teed2020raft} (estimating the flow between consecutive pairs of frames and chaining the flows to form trajectories), and a strong feature-matching method, DINO~\cite{caron2021emerging}. 

\begin{table*}[t!]
\centering
\resizebox{\textwidth}{!}{
\begin{tabular}{lC{1.5cm}C{1cm}C{1cm}C{1.5cm}C{1cm}C{1cm}C{1.5cm}C{1cm}C{1cm}C{1.5cm}}
  \toprule
  \multirow{2}*{Method} & \multirow{2}*{Training} & \multicolumn{3}{c}{\makecell[c]{PointOdyssey}} & \multicolumn{3}{c}{\makecell[c]{TAP-Vid-DAVIS~\cite{doersch2022tap}}} & \multicolumn{3}{c}{\makecell[c]{CroHD~\cite{voigtlaender2019mots}}}\\
 & & \small{MTE} $\downarrow$ & \small{$\delta \uparrow$} & \small{Survival} $\uparrow$& \small{MTE} $\downarrow$ & \small{$\delta \uparrow$} & \small{Survival} $\uparrow$& \small{MTE} $\downarrow$ & \small{$\delta \uparrow$} & \small{Survival} $\uparrow$\\
\midrule
TAP-Net~\cite{doersch2022tap}& \small{Pretrained}  & 92.00 & 23.75\% & 17.01\% & 10.56 & 53.40\% & 75.52\% & 101.12 & 23.39\% & 34.28\%\\
RAFT~\cite{teed2020raft}& \small{Pretrained} & 319.46 & 10.07\% & 32.61\% & 11.32 & 45.23\% & 75.39\% & 82.76 & 15.82\% & 62.22\%\\
DINO~\cite{caron2021emerging}& \small{Pretrained} & 118.38 & 8.61\% & 31.29\% & 24.57 & 33.05\% & 84.10\% & 116.80 & 8.46\% & 37.11\%\\
PIPs~\cite{harley2022particle}& \small{Pretrained}  & 147.45 & 16.53\% & 32.90\% & 6.30 & 58.22\% & 83.17\% & 19.23 & 40.23\% & 75.15\%\\
\midrule
TAP-Net~\cite{doersch2022tap}& \small{Kubric}  & 92.70 & 26.92\% & 9.59\% & 36.19 & 35.93\% & 69.19\% & 99.15 & 18.08\% & 28.43\%\\
TAP-Net~\cite{doersch2022tap}& \small{PointOdyssey} & 63.51& 28.37\% & 18.27\% & 25.93 & 41.73\% & 72.92\%&60.94 & 22.24\% & 35.00\%\\
PIPs~\cite{harley2022particle}& \small{PointOdyssey} & 63.98 & 27.34\% & 42.33\% & 5.14 &	61.33\% & 85.31\% & 11.94 & 44.02\% & 74.93\%\\
\midrule
PIPs+& \small{PointOdyssey} &  28.93 &  32.41\%&  49.88\% & 4.74 & 62.44\%& 88.39\% & \textbf{11.20} & \textbf{45.51}\%&  75.07\%\\
PIPs++& \small{PointOdyssey} & \textbf{26.95} & \textbf{33.64}\% & \textbf{50.47}\% & \textbf{4.60} & \textbf{63.45}\% & \textbf{88.42}\% & 11.21 & 44.09\% & \textbf{75.43}\% \\
\bottomrule
\end{tabular}
}
\vspace{-5pt}
\caption{Tracking performance on the PointOdyssey test set, TAP-Vid-DAVIS~\cite{doersch2022tap}, and CroHD~\cite{voigtlaender2019mots}. 
}
\vspace{-10pt}

\label{tab:exp}
\end{table*}
\vspace{0.5em}\noindent\textbf{Implementation details.} We use the official code of PIPs~\cite{harley2022particle}, RAFT~\cite{teed2020raft}, and DINO~\cite{teed2020raft}, and reimplement TAP-Net~\cite{doersch2022tap} in PyTorch. For RAFT and DINO, we use the pretrained weights for evaluation. We train and test PIPs and TAP-Net on our dataset, using 4-8 A5000 GPUs in parallel. In addition to evaluating on the PointOdyssey test set, we evaluate on TAP-Vid-DAVIS~\cite{doersch2022tap} and CroHD~\cite{sundararaman2021tracking}, which are real-video evaluation benchmarks. TAP-Vid-Davis mostly consists of videos of animals and humans, with sparse tracks annotated on foreground and background points; CroHD consists of surveillance-like recordings of crowds (\eg, in train stations), with tracks annotated on all human heads. We leave out TAP-Vid-Kinetics~\cite{doersch2022tap}, as it contains hard cuts, while our focus is on continuous video. 

\subsection{Evaluation}
\vspace{-0.25em}

\noindent\textbf{Evaluation metrics.} We report the average position accuracy $\delta_\textrm{avg}$ as proposed in TAP-Vid~\cite{doersch2022tap}. This measures the percentage of tracks within a threshold distance to ground truth, averaged over thresholds $\{1,2,4,8,16\}$, defined in a normalized resolution of $256 \times 256$.  
We use Median Trajectory Error (MTE) to measure the distance between the estimated tracks and ground truth tracks. While Harley et al.~\cite{harley2022particle} reported average trajectory error (ATE) using a \textit{mean}, we find the median more informative, as it is less sensitive to outliers. 
We also measure a ``Survival" rate, which we define as: the average number of frames until tracking failure, and report this as a ratio of video length. Failure is when L2 distance exceeds 50 pixels in the normalized $256\times256$ resolution. 

\begin{figure}[t]
\begin{center}
   \includegraphics[width=0.9\linewidth]{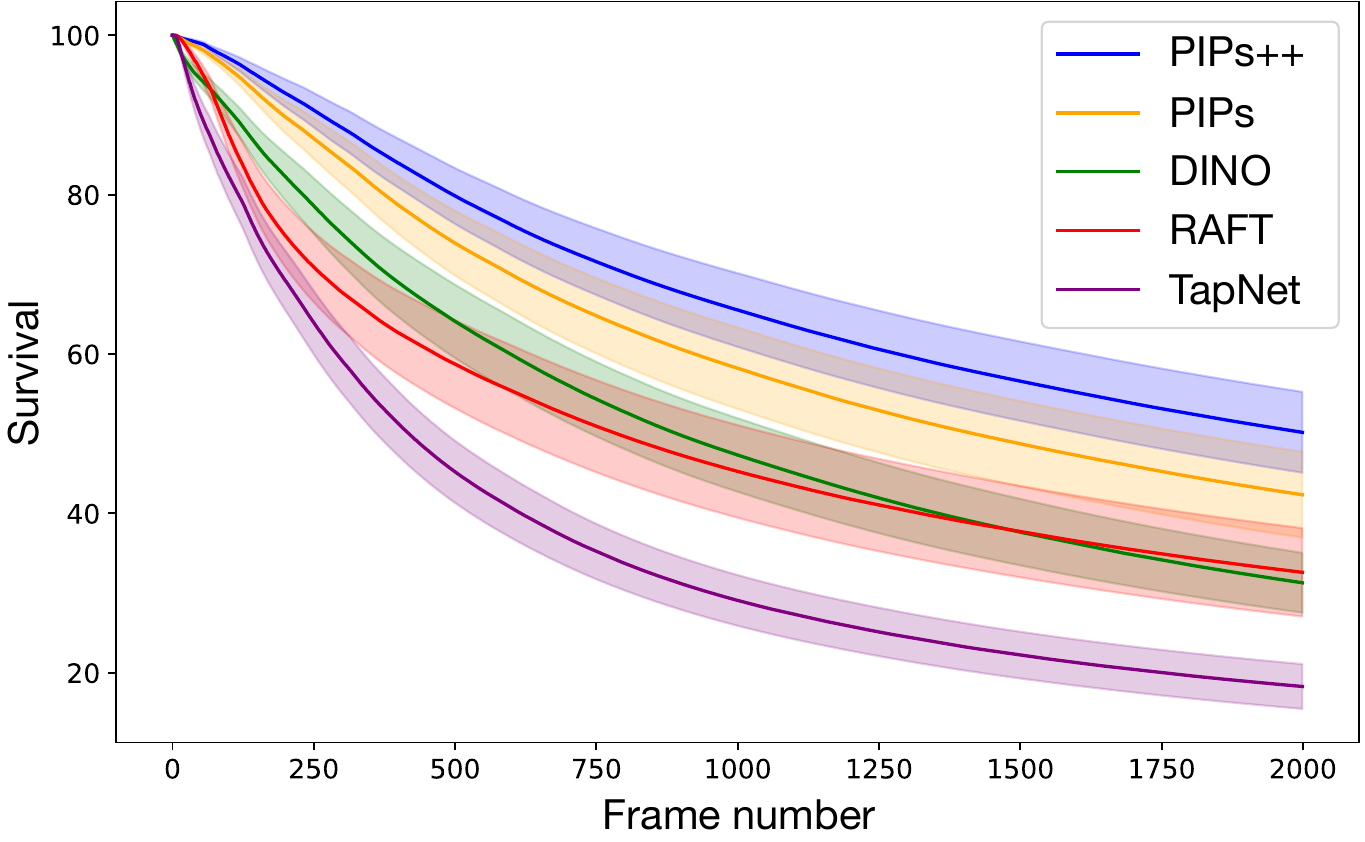}
\end{center}
\vspace{-10pt}
   \caption{Survival rate over time in PointOdyssey. Higher is better. We show the mean and standard deviation for each method. Note that the endpoints correspond to the values reported in Table 2.}
\vspace{-10pt}
\label{fig:survival}
\end{figure}

\vspace{0.5em}\noindent\textbf{Quantitative results.} We compile our results in~\cref{tab:exp}. 
First, inspecting results across rows (\ie, comparing methods), we can see that PIPs+ and PIPs++ achieve the best results among all methods, demonstrating the effectiveness of the wide temporal awareness. The narrow gap between PIPs+ and PIPs++ suggests that the multi-template strategy has only a modest effect, but is helpful on average. 
The results also demonstrate that prior methods perform better on real-world datasets when they are re-trained (from scratch) in our dataset. An exception here is TAP-Net~\cite{doersch2022tap}, where model trained by the authors (on Kubric) performs best; this is likely due to our smaller compute budget. All of our models are trained on 4-8 GPUs (c.f. 64 TPU-v3 cores in the original TAP-Net). 
Inspecting the results across columns (\ie, comparing datasets), we observe that PointOdyssey appears to be a more challenging benchmark than TAP-Vid-DAVIS~\cite{doersch2022tap} and CroHD~\cite{sundararaman2021tracking}.
We can also observe that the ranking of methods appears consistent among PointOdyssey and the two real-world datasets, suggesting a correlation between progress on PointOdyssey and progress on videos in the wild. \cref{fig:survival} plots survival rate over time in PointOdyssey, revealing that all methods struggle to keep tracks ``alive'' over long durations, but the PIPs models degrade more slowly than the rest. 

\begin{figure}[t!] 
    \centering
    \includegraphics[width=\linewidth]{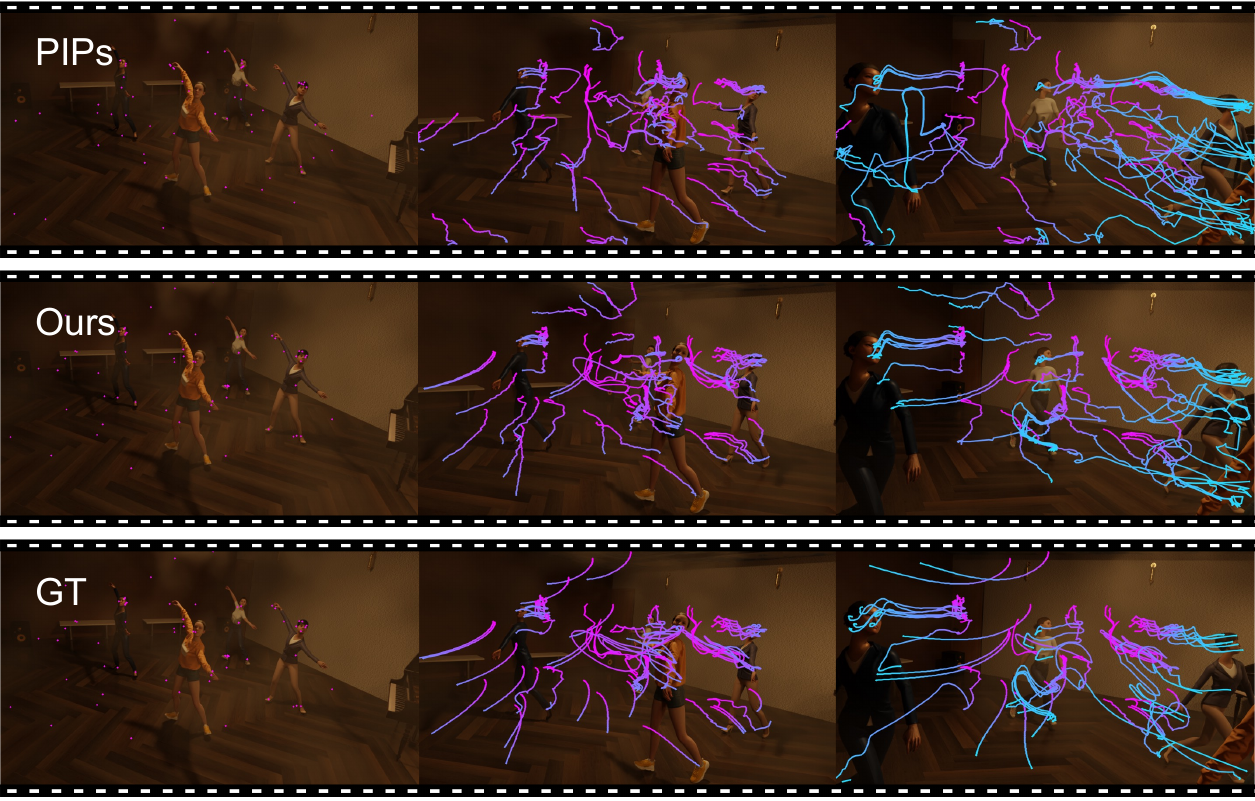}
    \vspace{-1em}
    \caption{Qualitative results on our dataset. Left to right columns show the start, middle and end frame respectively.}
    \label{fig:result}
    \vspace{-1em}
\end{figure}

\vspace{0.5em}\noindent\textbf{Qualitative results.} We show qualitative results in~\cref{fig:result}. Our method generates more-stable point trajectories compared to PIPs~\cite{harley2022particle} and other baselines. Please see the supplementary for video visualizations. We find that all methods have difficulty with targets which are close to boundaries (\eg, targets on thin objects are the hardest).

\section{Limitations}
\vspace{-0.25em}

PointOdyssey currently lacks large outdoor scenes where the camera travels a large distance, which is, for example, a frequent scenario in driving data~\cite{Dosovitskiy17}. It would be interesting to explore long-range agent-scene and agent-agent interactions in that context. We also note that our human and animal motion profiles are limited by our base datasets, and this constraint could be lifted with the help of recent generative models~\cite{rempe2021humor, yi2022mime}.
While the focus in this paper is on point tracking, our dataset connects to a wide range of applications which we have not yet explored, such as 3D scene flow estimation, novel view synthesis in dynamic scenes (which would be especially challenging with PointOdyssey's atmospheric effects), human and animal pose estimation, and ego-centric vision. 
Finally, while PIPs++ takes a step toward modelling longer-range temporal priors, it is still a fairly low-level tracker, relying entirely on appearance-matching cues and a temporal prior. This leaves open the challenge of leveraging scene-level and semantic cues for tracking, where we expect PointOdyssey's training data will be especially valuable.

\section{Conclusion}
\vspace{-0.25em}
PointOdyssey is a large-scale synthetic dataset, and data generator, for long-term point tracking. The data is diverse and naturalistic, making it an ideal resource for training general-purpose fine-grained trackers. We demonstrate its usefulness through a new tracker called PIPs++, which leverages long-term temporal context and outperforms state-of-the-art. PointOdyssey also opens opportunities for developing trackers which utilize scene-level and semantic cues, though we have not explored this yet. We hope our work will also be useful beyond point tracking, enabling work in 3D and 4D scene analysis, and higher-level video understanding. 

\vspace{0.5em}\noindent{\textbf{Acknowledgments.}}
The authors thank Andrew Zisserman for feedback on an early version of the title. This work was supported by the Toyota Research Institute under the University 2.0 program, ARL grant W911NF-21-2-0104, and a Vannevar Bush Faculty Fellowship. 

{\small
\bibliographystyle{ieee_fullname}
\bibliography{99_refs}
}

\clearpage
\appendix
\section*{Supplementary Material}

\section{Implementation Details}

\noindent\textbf{Training Loss.}
We train PIPs+ and PIPs++ using the weighted $L_1$ distance between the estimated trajectory and the ground truth trajectory across iterative updates as proposed by Harley et al.~\cite{harley2022particle}. For any query point $p_n$, we compute the loss as follows:
\begin{equation}
    \mathcal{L}_n = \sum_{k=1}^K \left( \gamma^{K-k}\frac{1}{T}\sum_{t=1}^T||p^k_{t,n}-p^*_{t,n}||_1 \right)\,,
\end{equation}
where $p^k_{t,n}$ denotes the estimated position at timestep $t$, from iteration $k$, and $p^*_{t,n}$ is the ground truth. We set $\gamma=0.8$ in our experiment. The full training loss is obtained by averaging the per-point loss across all the N query points:
\begin{equation}
    \mathcal{L} = \frac{1}{N}\sum_{n=1}^N\mathcal{L}_n\,.
    \label{eq: loss}
\end{equation}
The loss is applied even when the target is occluded, which asks the model to estimate the track during visibility gaps. Note that PIPs~\cite{harley2022particle} is also trained with a visibility classification loss, to predict whether a point is occluded, along with a score loss that supervises the similarity score map to peak at the correct locations to help the network converge faster. We find that those losses can be omitted without harming tracking performance, and therefore we only use the $\mathcal{L}$ loss presented in~\cref{eq: loss}.

\vspace{0.5em}\noindent\textbf{Test Time Trajectory Chaining.}
In order to track with PIPs for more than 8 frames, Harley et al.~\cite{harley2022particle} link multiple 8-frame predictions. The linking strategy works per-point: after tracking for 8 frames and estimating visibility on each frame, the tracker is re-initialized on the last timestep whose visibility exceeds a threshold. While PIPs runs quickly within 8-frame clips, this chaining strategy leads to an overall FPS of 3.6 (at $720\times1080$ on an Nvidia V100 GPU). 
Although PIPs++ does not in principle require a linking strategy, our GPU memory constraints necessitate one. 
We use a very simple strategy: we predict 36 timesteps at a time, in sliding-window fashion, with \textit{no overlap} between the windows. Since there is no visibility check, this is very fast, leading to an overall FPS of 55.2. The model can be run with larger or smaller windows, but in a small grid search we found that 36 works best, possibly because this was also the sequence length used at training time. 

\section{Additional Visualizations of PointOdyssey}

\begin{figure*}[t!] 
    \centering
    \vspace{5pt}
    \includegraphics[width=\linewidth]{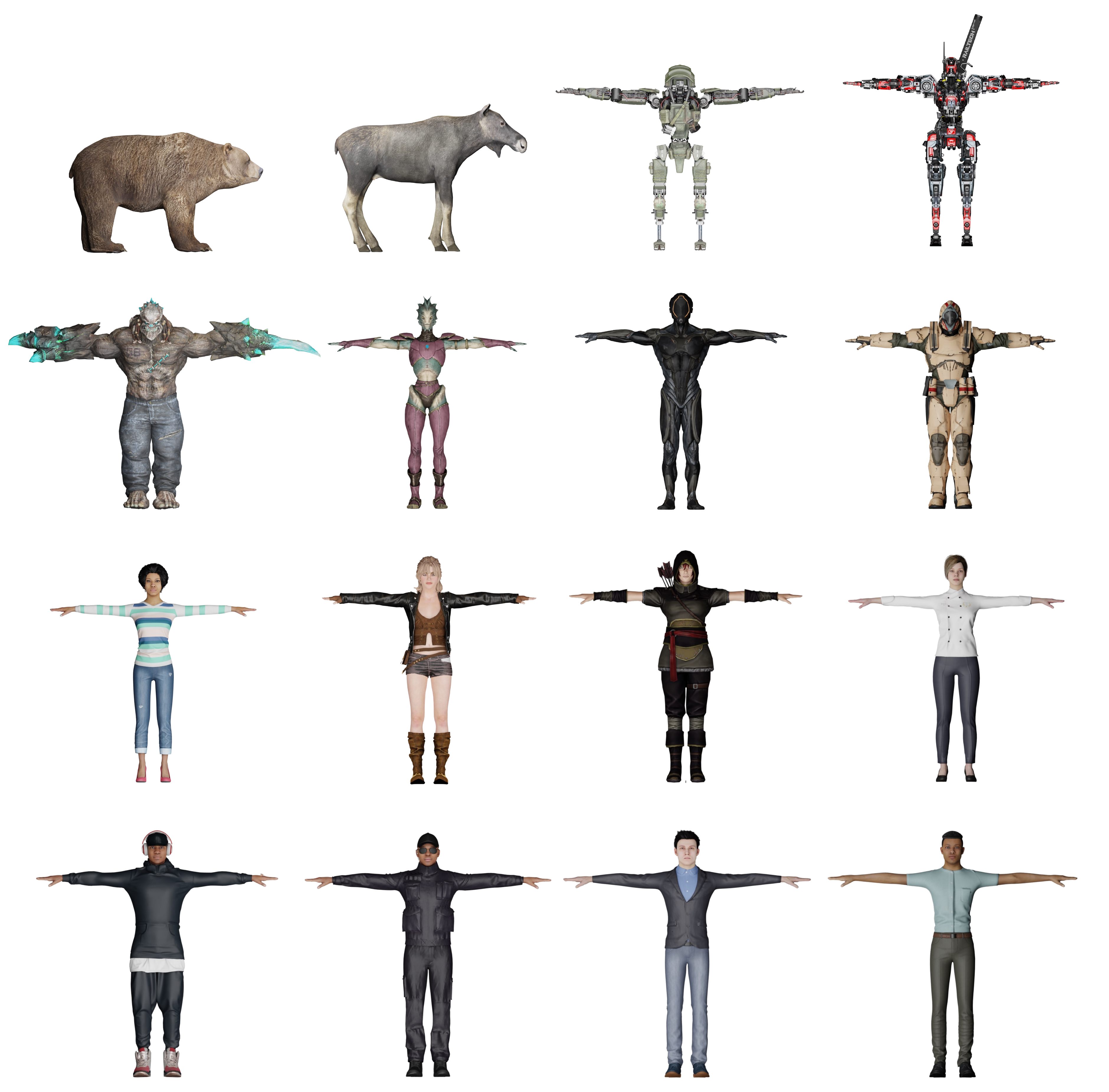}
    \caption{Sample characters from our dataset.} 
    \vspace{5pt}
    \label{fig:supp_chara}
\end{figure*}

\begin{figure*}[t!] 
    \centering
    \vspace{-10pt}
    \includegraphics[width=0.9\linewidth]{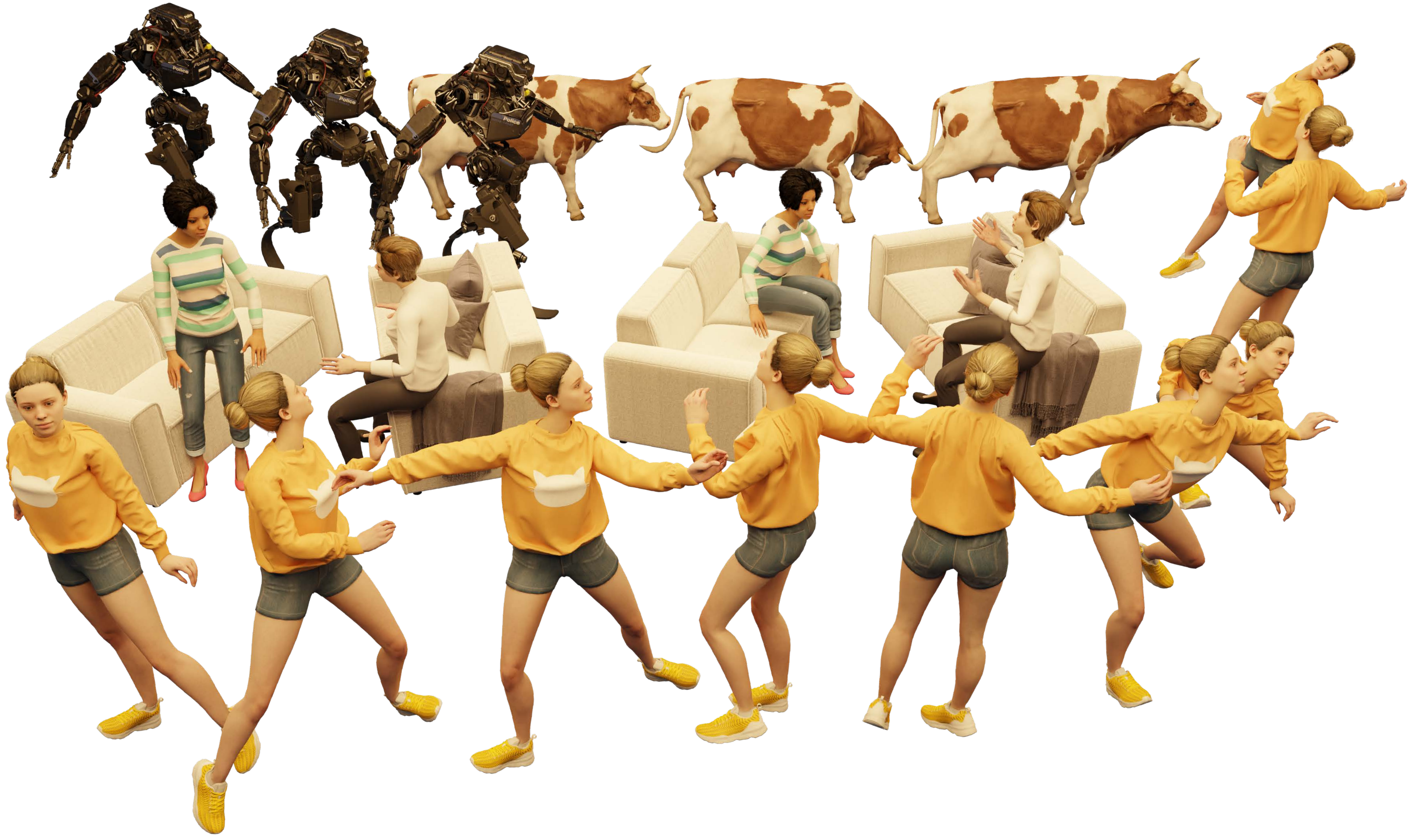}
    \vspace{-10pt}
    \caption{Sample motions from our dataset.} 
    \label{fig:supp_motion}
\end{figure*}

Sample animated characters from the dataset are shown in ~\cref{fig:supp_chara}. 
By retargeting motion data to these characters, we are able to generate a wide range of interacting sequences, as illustrated in \cref{fig:supp_motion}. 

\begin{figure*}[t!] 
    \centering
    \includegraphics[width=0.9\linewidth]{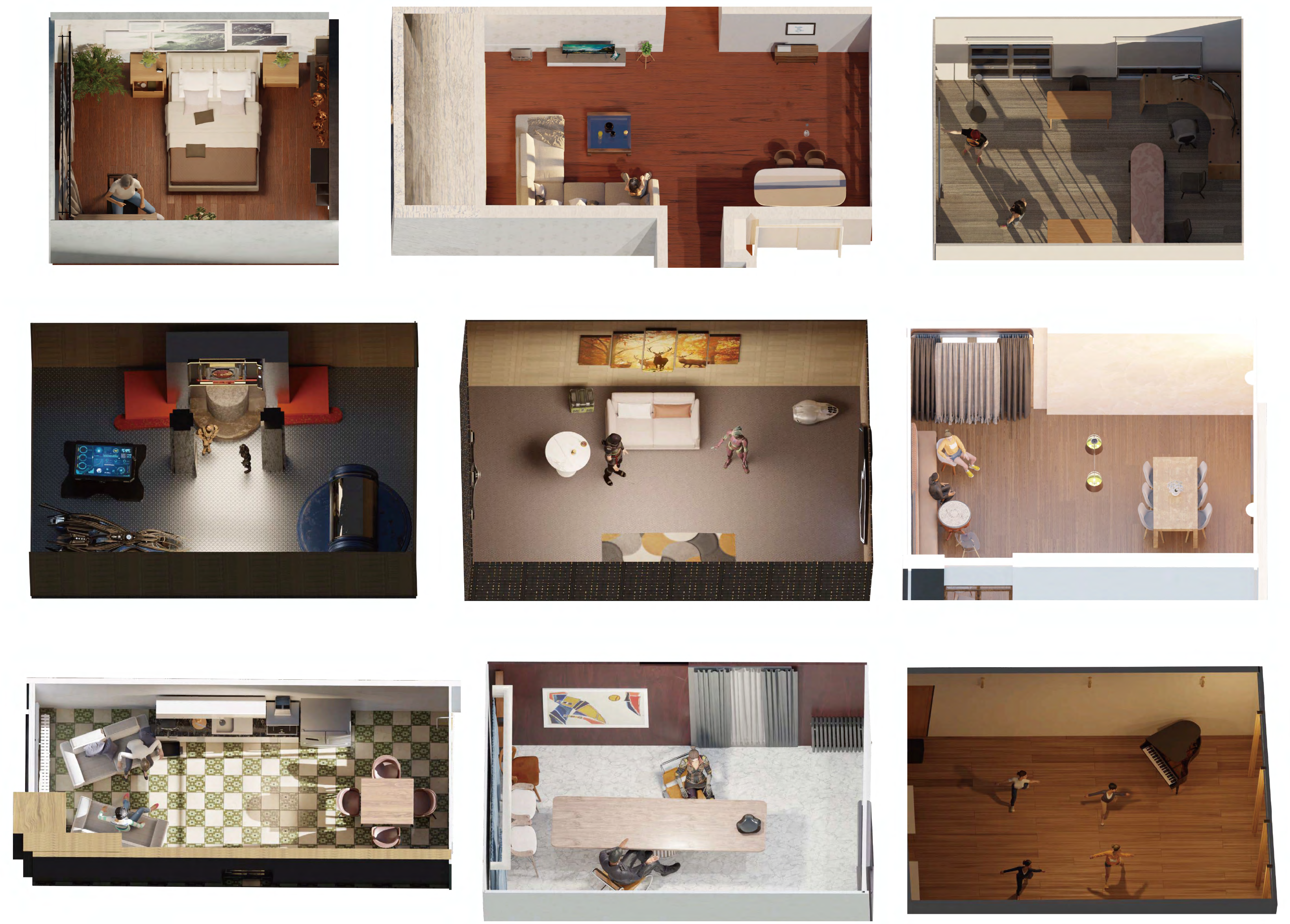}
    \caption{Examples of the rebuilt 3D scenes, with humans at a random timestep in their trajectories.} 
    \label{fig:supp_scene}
\end{figure*}

Samples of our re-built motion capture environments are shown in \cref{fig:supp_scene}. 

\begin{figure*}[t!] 
    \centering
    \vspace{20pt}
    \includegraphics[width=\linewidth]{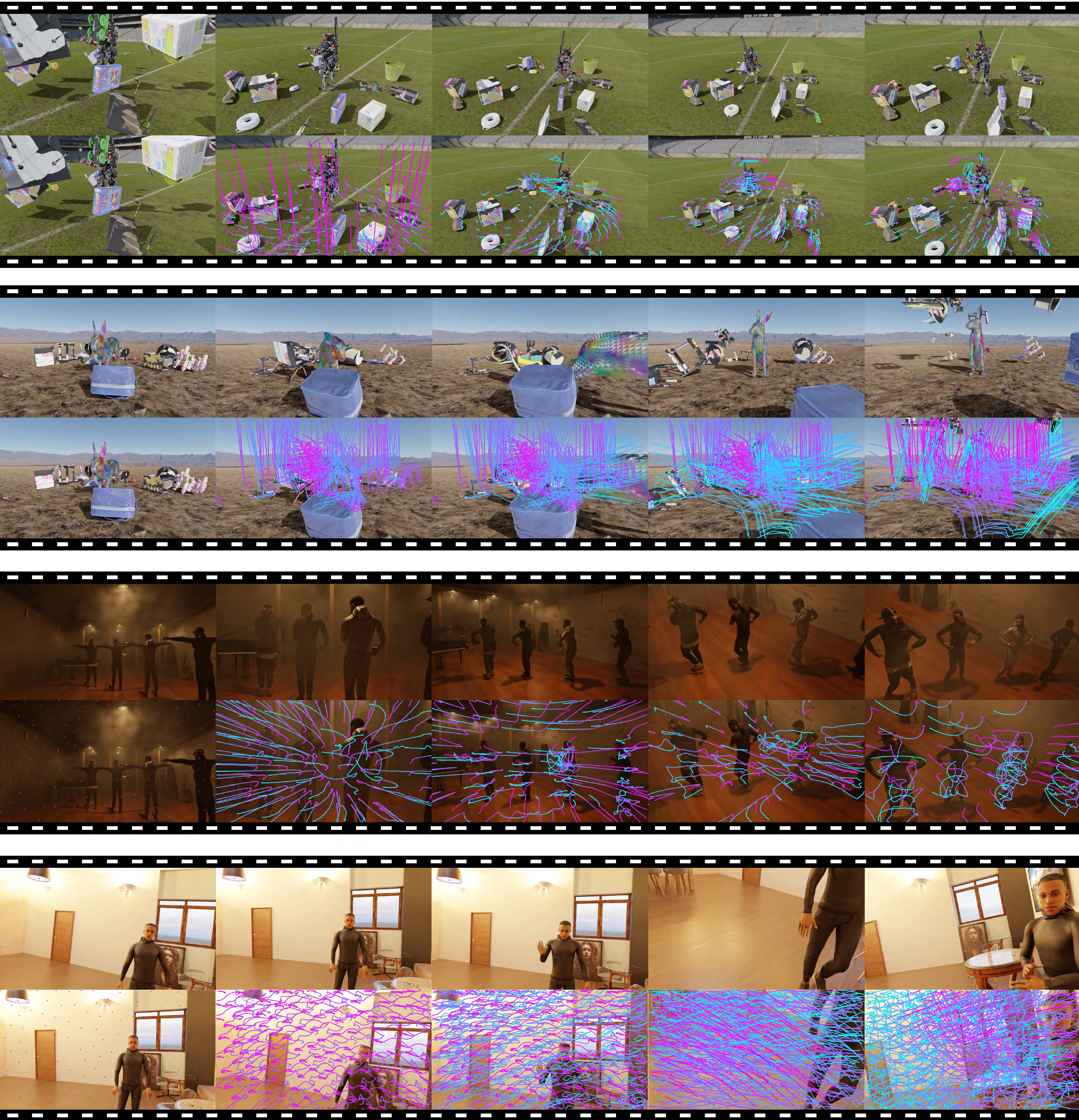}
    \caption{Sample RGB videos and point trajectories from PointOdyssey. The four rows show: (1) a robot walking in a stadium interacting with random objects; (2) a crystal rabbit in the desert interacting with random objects; (3) four human characters dancing in a room; (4) an ego-centric view of one character talking to another.} 
    \label{fig:supp_seq}
    \vspace{30pt}
\end{figure*}

~\cref{fig:supp_seq} shows sample images pixel trajectories from the dataset. 

We recommend watching the supplementary video for additional visualizations. 

\section{Additional Results}

\begin{figure*}[t!] 
    \centering
    \includegraphics[width=\linewidth]{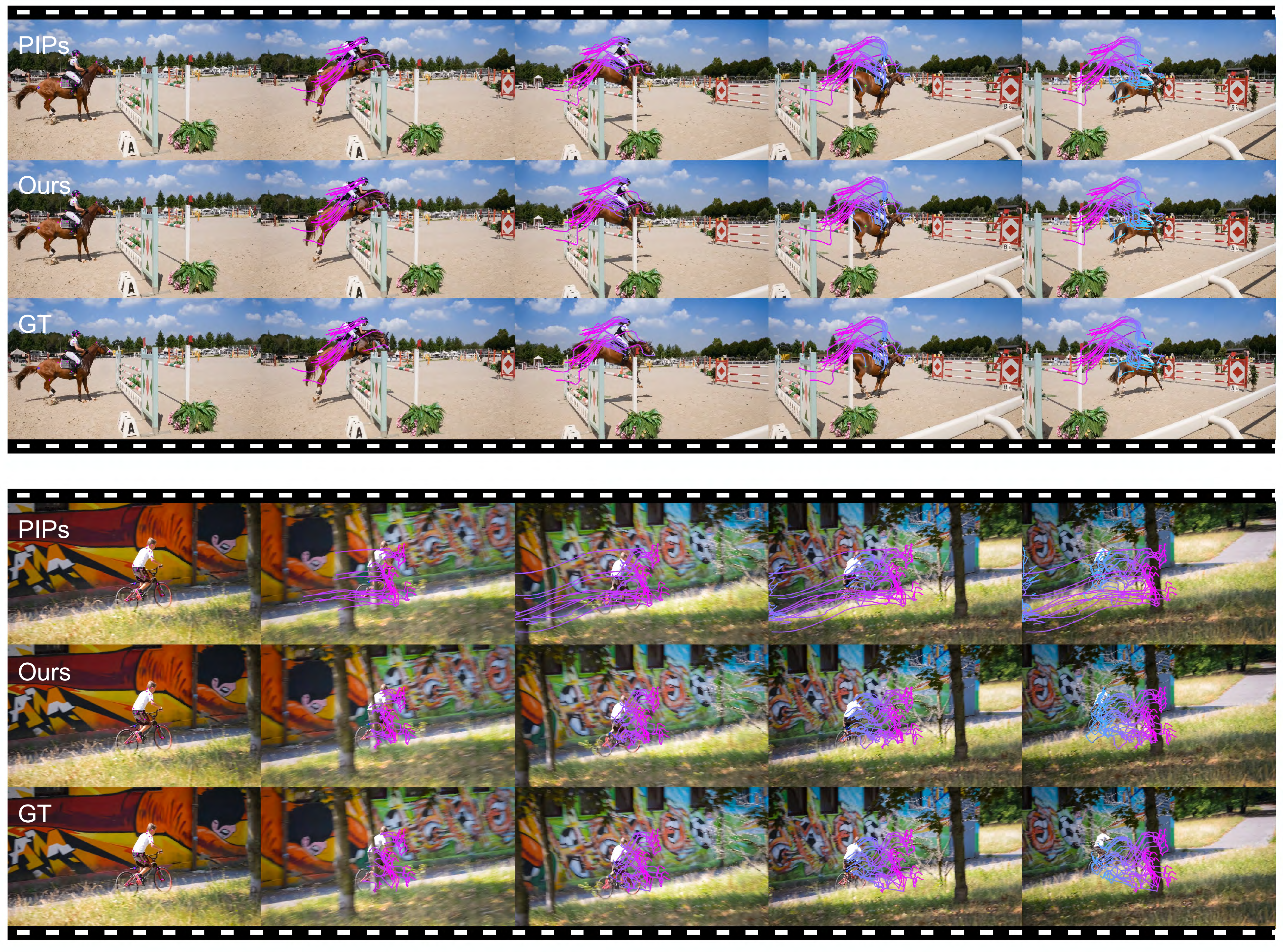}
    \caption{Qualitative results on Tap-Vid-DAVIS~\cite{doersch2022tap}. While PIPs~\cite{harley2022particle} can track well during high visibility (as shown in the top horse riding sequence), it becomes unreliable when occlusions occur (as shown in the bike riding sequence). PIPs++ can withstand such occlusions more frequently than PIPs, likely thanks to its wider temporal receptive field.} 
    \label{fig:supp_results}
\end{figure*}

We show the performance of PIPs~\cite{harley2022particle} and our PIPs++ method on real-world data in~\cref{fig:supp_results}. While PIPs can track visible points effectively, it struggles with occlusions. Trajectories from PIPs++ are on average less sensitive to occlusions. 


\end{document}